\documentclass[11pt]{article}

\usepackage[utf8]{inputenc}
\usepackage[T1]{fontenc}
\usepackage{amsmath,amssymb,amsfonts}
\usepackage{graphicx}
\graphicspath{{./figures/}}
\usepackage{booktabs}
\usepackage{xcolor}
\usepackage[margin=1in]{geometry}
\usepackage{natbib}
\usepackage{multirow}
\usepackage{float}
\usepackage[section]{placeins}
\usepackage{textcomp}
\usepackage{hyperref}
\usepackage{enumitem}

\title{Time Series Foundation Models for Energy Load Forecasting on Consumer Hardware:\\A Multi-Dimensional Zero-Shot Benchmark}

\author{
  Luigi Simeone\\
  Independent Researcher, Turin, Italy\\
  \texttt{luigi.simeone@tech-management.net}
}

\date{February 2026}

\begin{document}

\maketitle

\begin{abstract}
Time Series Foundation Models (TSFMs) have introduced zero-shot prediction capabilities that bypass the need for task-specific training. Whether these capabilities translate to mission-critical applications such as electricity demand forecasting---where accuracy, calibration, and robustness directly affect grid operations---remains an open question.

We present a multi-dimensional benchmark evaluating four TSFMs (Chronos-Bolt, Chronos-2, Moirai-2, and TinyTimeMixer) alongside Prophet as an industry-standard baseline and two statistical references (SARIMA and Seasonal Naive), using ERCOT hourly load data from 2020 to 2024. All experiments run on consumer-grade hardware (AMD Ryzen~7, 16\,GB RAM, no GPU). The evaluation spans four axes: (1)~context length sensitivity from 24 to 2048~hours, (2)~probabilistic forecast calibration, (3)~robustness under distribution shifts including COVID-19 lockdowns and Winter Storm Uri, and (4)~prescriptive analytics for operational decision support.

The top-performing foundation models achieve MASE values near 0.31 at long context lengths ($C = 2048$\,h, day-ahead horizon), a 47\% reduction over the Seasonal Naive baseline. The inclusion of Prophet exposes a structural advantage of pre-trained models: Prophet fails when the fitting window is shorter than its seasonality period (MASE $>$~74 at 24-hour context), while TSFMs maintain stable accuracy even with minimal context because they recognise temporal patterns learned during pre-training rather than estimating them from scratch. Calibration varies substantially across models---Chronos-2 produces well-calibrated prediction intervals (95\% empirical coverage at 90\% nominal level) while both Moirai-2 and Prophet exhibit overconfidence ($\sim$70\% coverage). We provide practical model selection guidelines and release the complete benchmark framework for reproducibility.
\end{abstract}


%==============================================================================
\section{Introduction}
%==============================================================================

Electricity demand forecasting is central to power system operations. Day-ahead and week-ahead load predictions determine generation scheduling, demand response activation, and reserve procurement. Forecast errors carry immediate economic consequences: underestimation triggers expensive peaking generation or, in the worst case, load shedding; overestimation leads to unnecessary fuel consumption and elevated operating costs \citep{haben2021review, hong2016probabilistic}. In deregulated markets such as ERCOT, where reserve margins have thinned to single-digit percentages, the cost of a 1\% improvement in forecast accuracy can translate to millions of dollars in annual savings across the system \citep{hong2010short}.

Traditional forecasting methods span a wide range of complexity. Statistical approaches---ARIMA, exponential smoothing, and their seasonal variants---remain the workhorse of many grid operators \citep{taylor2010triple, hyndman2021forecasting}. Machine learning methods, from gradient-boosted trees to deep neural networks such as N-BEATS \citep{oreshkin2020nbeats} and Temporal Fusion Transformers \citep{lim2021temporal}, have demonstrated improved accuracy, particularly when rich exogenous features (weather, calendar, and price signals) are available. Prophet \citep{taylor2018forecasting}, an additive decomposition model originally released as Facebook Prophet, has become a de facto industry standard for many operational teams due to its ease of deployment and interpretable components.

All of these methods share a common requirement: they must be fitted on historical data from the target series before producing forecasts. This creates practical friction in several scenarios. Deploying forecasting to a new substation or microgrid requires collecting sufficient local history. Concept drift---caused by electrification trends, new renewable capacity, or behavioural shifts such as COVID-19 lockdowns---necessitates periodic retraining. The fitting procedure itself introduces computational overhead that scales with data length and model complexity.

Time Series Foundation Models (TSFMs) offer a distinct alternative. Pre-trained on large corpora of heterogeneous time series data, these models generate forecasts for previously unseen series without any task-specific fitting---a capability referred to as zero-shot forecasting. Recent architectures include Amazon's Chronos family \citep{ansari2024chronos, ansari2025chronos2}, which tokenises continuous values and applies encoder-decoder and encoder-only transformers; Salesforce's Moirai \citep{woo2024unified, liu2025moirai2}, a universal time series forecasting transformer family trained on large-scale multi-domain corpora; IBM's TinyTimeMixer (TTM) \citep{ekambaram2024tiny}, an MLP-Mixer architecture with fewer than one million parameters; Google's TimesFM \citep{das2024decoder}, a decoder-only model; and Lag-Llama \citep{rasul2023lag}, which adapts the LLaMA architecture for probabilistic forecasting.

While TSFMs have shown competitive results on standard benchmarking suites, their suitability for energy forecasting remains insufficiently characterised along several dimensions that matter for operational deployment:

\begin{enumerate}[label=\textbf{RQ\arabic*:},leftmargin=2cm]
    \item How much historical context do TSFMs need to match or exceed fitted baselines? Energy demand exhibits multi-seasonal patterns (daily, weekly, annual) that may require extended context windows.
    \item Are the probabilistic forecasts produced by TSFMs well-calibrated? Grid operators rely on prediction intervals for reserve planning and risk management; overconfident intervals can lead to insufficient reserves.
    \item How robust are TSFM predictions under distribution shift? Demand patterns change during extreme weather events, pandemics, and policy interventions.
    \item Can TSFM forecasts support prescriptive analytics? Beyond point accuracy, do probabilistic outputs enable actionable decisions such as demand response scheduling and storage optimisation?
\end{enumerate}

This paper addresses these questions through a systematic benchmark on ERCOT hourly load data spanning January 2020 through December 2024, executed entirely on consumer-grade hardware (AMD Ryzen~7 8845HS, 16\,GB RAM, no dedicated GPU). The hardware constraint is deliberate: it evaluates whether TSFMs are accessible to utilities, cooperatives, and researchers who lack cloud or high-performance computing infrastructure.

\subsection{Contributions}

This work makes five contributions:

\begin{enumerate}
    \item A \textbf{multi-dimensional zero-shot TSFM benchmark for energy forecasting} evaluating four foundation models across accuracy, calibration, robustness, and prescriptive utility, with over 2{,}300 individual forecast evaluations.
    \item A \textbf{context length sensitivity analysis} spanning 24 to 2048~hours, revealing that TSFMs benefit from extended context without performance saturation, in contrast to fitted baselines that fail at short context lengths.
    \item An \textbf{uncertainty calibration assessment} exposing substantial inter-model differences---from near-perfect calibration (Chronos-2) to problematic overconfidence (Moirai-2, Prophet) and vacuous uncertainty (TTM).
    \item A \textbf{comparison with Prophet as an industry baseline}, demonstrating that Prophet's reliance on local parameter estimation causes catastrophic failure with limited context, whereas TSFMs leverage pre-trained pattern recognition to maintain accuracy.
    \item A \textbf{prescriptive analytics framework} quantifying the operational value of probabilistic forecasts for demand response scheduling, reserve planning, and storage optimisation.
\end{enumerate}

%==============================================================================
\section{Related Work}
%==============================================================================

\subsection{Time Series Foundation Models}

The application of the foundation model paradigm to time series data has produced a rapidly growing family of architectures. These can be grouped by their approach to the core challenge of representing continuous-valued temporal sequences.

\textbf{Tokenisation-based approaches.} Chronos \citep{ansari2024chronos} pioneered the idea of treating time series forecasting as a language modelling problem: continuous values are quantised into discrete tokens via scaling and binning, then processed by a T5-style encoder-decoder transformer. The Chronos family has since expanded to include Chronos-Bolt (optimised for efficiency) and Chronos-2 \citep{ansari2025chronos2}, an encoder-only model with a group attention mechanism supporting multivariate inputs and covariates. TimesFM \citep{das2024decoder} takes a decoder-only approach, pre-training a large transformer on Google's internal time series corpus.

\textbf{Continuous-value approaches.} Moirai \citep{woo2024unified} processes raw continuous values through a universal transformer with any-variate attention, trained on the LOTSA dataset comprising over 27 billion observations across diverse domains. Moirai~2.0 \citep{liu2025moirai2} replaces the masked-encoder architecture with a decoder-only design employing quantile loss and multi-token prediction, trained on a larger corpus of 36 million series ($\sim$295 billion observations). Lag-Llama \citep{rasul2023lag} adapts the LLaMA architecture for probabilistic time series forecasting using lagged features as inputs, producing distributional outputs via a Student-$t$ likelihood.

\textbf{Lightweight architectures.} TinyTimeMixer (TTM) \citep{ekambaram2024tiny} departs from the transformer paradigm entirely, using an MLP-Mixer architecture that processes patched time series through channel-mixing and time-mixing layers. With fewer than one million parameters, TTM is designed for edge deployment and few-shot adaptation.

\textbf{Patch-based transformers.} PatchTST \citep{nie2023patchtst} segments time series into subseries-level patches, applies a vanilla transformer encoder, and demonstrated that patch-level tokenisation can outperform point-level approaches. iTransformer \citep{liu2024itransformer} inverts the standard transformer architecture by applying attention across variates rather than time steps, treating each variate's temporal sequence as a token.

Prior benchmarking studies have typically evaluated TSFMs on heterogeneous collections of standard datasets (Monash, M4, ETTh). Domain-specific evaluations remain scarce, particularly for energy applications where calibration and robustness matter as much as point accuracy.

\subsection{Energy Load Forecasting}

Short-term load forecasting (STLF) has a long research history. Statistical methods remain competitive: Taylor \citep{taylor2010triple} demonstrated that double and triple seasonal exponential smoothing methods capture the multi-periodic structure of electricity demand effectively. Hong et al.\ \citep{hong2016probabilistic} established probabilistic forecasting as a priority for the energy sector, arguing that point forecasts alone are insufficient for grid operations requiring risk quantification.

Deep learning approaches have progressively entered the field. N-BEATS \citep{oreshkin2020nbeats} introduced a pure neural architecture with interpretable basis expansion, achieving strong performance without requiring feature engineering. The Temporal Fusion Transformer (TFT) \citep{lim2021temporal} combined multi-horizon forecasting with variable selection and temporal attention, becoming a reference architecture for multi-step energy prediction. DeepAR \citep{salinas2020deepar} extended autoregressive recurrent networks to produce full predictive distributions.

Prophet \citep{taylor2018forecasting} occupies a distinct position in this landscape. Designed for business time series with strong seasonalities and trend changes, Prophet decomposes a time series into trend, seasonal (Fourier-based), and holiday components. Its popularity in operational energy forecasting stems from its ease of use, automatic seasonality detection, and built-in uncertainty quantification. However, Prophet must estimate all of its parameters---trend changepoints, Fourier coefficients for daily and weekly seasonality, and observation noise---from the provided fitting data. As we show in this study, this creates a structural vulnerability when fitting data is limited.

\subsection{Probabilistic Forecast Evaluation}

Proper evaluation of probabilistic forecasts requires scoring rules that jointly assess calibration and sharpness. The Continuous Ranked Probability Score (CRPS) \citep{gneiting2007strictly} is a strictly proper scoring rule that measures the compatibility of the full predictive distribution with observed outcomes. For interval forecasts, the Winkler Score \citep{winkler1972scoring} penalises both interval width and coverage violations. Kuleshov et al.\ \citep{kuleshov2018calibrated} demonstrated that many neural network-based forecasters produce systematically miscalibrated uncertainty estimates, motivating post-hoc recalibration methods. For energy applications, Pinson \citep{pinson2013wind} established that well-calibrated probabilistic forecasts are essential for reserve planning and market participation in renewable-dominated grids.

%==============================================================================
\section{Experimental Methodology}
%==============================================================================

\subsection{Dataset}

We use ERCOT hourly system-wide electricity demand data from January 2020 to December 2024, obtained via the EIA Open Data API. The dataset comprises 43{,}732 observations measured in megawatts (MW). ERCOT manages the electrical grid for approximately 90\% of Texas's electric load, serving over 26 million customers. The time series exhibits three dominant periodicities: a 24-hour diurnal cycle driven by human activity and solar radiation, a 168-hour weekly cycle reflecting weekday/weekend demand differences, and an annual cycle with summer peaks (air conditioning load, reaching $\sim$75{,}000\,MW) and spring/autumn troughs ($\sim$25{,}000\,MW).

The study period encompasses three operationally distinct regimes. Normal operations (2022--2024) provide a baseline with typical seasonal patterns. The COVID-19 lockdown period (March--April 2020) introduced a 15--20\% demand reduction with altered diurnal patterns as commercial activity shifted to residential settings. Winter Storm Uri (February 2021) caused a cascading grid failure with demand first spiking due to extreme heating load and then collapsing as load shedding was implemented---an event without precedent in the ERCOT system.

\subsection{Models}

We evaluate seven models spanning three categories: pre-trained foundation models (zero-shot), an industry-standard fitted baseline (Prophet), and statistical baselines.

\textbf{Foundation Models (zero-shot inference, no fitting):}

\textit{Chronos-Bolt} \citep{ansari2024chronos} uses a T5-based encoder-decoder transformer with patch-based input tokenisation. Unlike the original Chronos, which samples from a learned token distribution via cross-entropy, Chronos-Bolt directly produces quantile forecasts (9 quantiles from 0.1 to 0.9), achieving approximately 250$\times$ faster inference. We use the \textit{small} variant (\texttt{amazon/chronos-bolt-small}, $\sim$48M parameters).

\textit{Chronos-2} \citep{ansari2025chronos2} employs an encoder-only architecture that alternates between time attention (across patches within a single series) and group attention (across related series at each patch index), enabling in-context learning for univariate, multivariate, and covariate-informed forecasting. Here we use it in univariate zero-shot mode for consistency with the benchmark (\texttt{amazon/chronos-2}, $\sim$120M parameters).

\textit{Moirai-2} \citep{liu2025moirai2} is a decoder-only universal time series forecasting transformer that employs quantile forecasting with multi-token prediction. It is pre-trained on a curated corpus of 36 million time series comprising approximately 295 billion observations, including subsets of the LOTSA dataset \citep{woo2024unified}, KernelSynth data, and anonymised operational telemetry. Compared to Moirai~1.0, Moirai-2 replaces masked-encoder training and mixture-distribution outputs with a simpler decoder-only architecture and quantile loss. We use the \textit{small} variant (\texttt{Salesforce/moirai-2.0-R-small}, $\sim$11M parameters).

\textit{TinyTimeMixer (TTM)} \citep{ekambaram2024tiny} replaces the transformer with an MLP-Mixer operating on patched time series segments. With fewer than 1M parameters, it is over an order of magnitude smaller than the other foundation models and is designed for scenarios where computational resources are limited. We use the \textit{r2} variant (\texttt{ibm-granite/granite-timeseries-ttm-r2}).

\textbf{Industry Baseline (fitted at each invocation):}

\textit{Prophet} \citep{taylor2018forecasting} decomposes the time series into trend, weekly seasonality, and daily seasonality components using Fourier series. We configure Prophet with multiplicative seasonality mode and \texttt{interval\_width=0.90} to produce 90\% prediction intervals. At each forecast, Prophet fits its full parameter set---trend changepoints, Fourier coefficients, and noise variance---on the provided context window. This design choice is central to our analysis: unlike foundation models, Prophet has no pre-trained knowledge and must learn all temporal patterns from scratch at every invocation.

\textbf{Statistical Baselines:}

\textit{SARIMA} is specified as ARIMA$(2,1,2)(1,1,1)_{24}$ following standard Box-Jenkins methodology with seasonal differencing at the 24-hour period. Like Prophet, SARIMA fits on the context window at each invocation.

\textit{Seasonal Naive} repeats observed values from exactly 168~hours (one week) prior. This baseline defines the MASE denominator and represents the ``no skill'' reference: MASE values above 1.0 indicate performance worse than simply repeating last week's pattern.

\subsubsection{Computational Constraints and Model Selection Rationale}

All experiments are executed on a single consumer laptop equipped with an AMD Ryzen~7 8845HS processor and 16\,GB of system RAM, without any dedicated GPU. This constraint is not incidental but constitutes a deliberate design choice. Many utilities, rural cooperatives, and academic researchers lack access to GPU clusters. Evaluating TSFMs under these conditions tests whether their advertised zero-shot capabilities remain practical outside well-resourced environments.

The hardware constraint also determined model selection. TimesFM \citep{das2024decoder} requires GPU acceleration for inference within reasonable time. Lag-Llama \citep{rasul2023lag} exceeds available memory at longer context lengths. TabPFN is not designed for univariate temporal forecasting. The four selected TSFMs (Chronos-Bolt, Chronos-2, Moirai-2, TTM) represent the models that execute successfully on the target hardware across all tested context lengths, spanning diverse architectural families: encoder-decoder transformer, encoder-only transformer, decoder-only transformer, and MLP-Mixer.

\subsection{Evaluation Protocol}

We test eight context lengths $C \in \{24, 48, 96, 168, 336, 512, 1024, 2048\}$ hours, ranging from one day to approximately twelve weeks. This range is chosen to span from sub-daily context (well below the weekly seasonality period) to multi-month context that captures multiple seasonal cycles. Forecast horizons are $H \in \{24, 168\}$~hours, corresponding to day-ahead and week-ahead predictions---the two horizons most relevant to grid operations.

Evaluation uses three non-overlapping test periods representing distinct operational regimes: Summer 2023 (July--August, high demand with stable patterns), Winter 2022--2023 (December--February, heating load with holiday effects), and COVID-19 (March--April 2020, unprecedented demand reduction). Within each period, we extract seven non-overlapping rolling windows to reduce variance from window placement. The total number of individual forecast evaluations is $7 \times 8 \times 2 \times 3 \times 7 = 2{,}352$ across all model-context-horizon-period-window combinations.

\subsection{Metrics}

We employ four complementary metrics spanning point accuracy, distributional quality, calibration, and interval sharpness.

\textbf{Mean Absolute Scaled Error (MASE)} normalises the mean absolute forecast error by the in-sample seasonal naive error with period $m = 168$ (weekly):
\begin{equation}
\text{MASE} = \frac{\frac{1}{H}\sum_{h=1}^{H}|y_{t+h} - \hat{y}_{t+h}|}{\frac{1}{T-m}\sum_{t=m+1}^{T}|y_t - y_{t-m}|}
\label{eq:mase}
\end{equation}
MASE below 1.0 indicates improvement over the Seasonal Naive baseline. The choice of $m = 168$ reflects the dominant weekly periodicity in electricity demand.

\textbf{Continuous Ranked Probability Score (CRPS)} evaluates the full predictive distribution:
\begin{equation}
\text{CRPS}(F, y) = \int_{-\infty}^{\infty} \left[F(x) - \mathbf{1}(x \geq y)\right]^2 dx
\label{eq:crps}
\end{equation}
where $F$ is the predictive CDF and $y$ the observed value. CRPS is a strictly proper scoring rule \citep{gneiting2007strictly}, meaning it is minimised only when the predictive distribution matches the true data-generating distribution.

\textbf{Empirical Coverage} at nominal level $\alpha$ measures the proportion of observations falling within the $[(\frac{1-\alpha}{2}), (\frac{1+\alpha}{2})]$ quantile interval. For a well-calibrated model at the 90\% level, empirical coverage should be close to 90\%.

\textbf{Winkler Score} \citep{winkler1972scoring} at level $\alpha$ penalises both interval width and coverage violations:
\begin{equation}
W_\alpha = \begin{cases}
(u - l) & \text{if } l \leq y \leq u \\
(u - l) + \frac{2}{\alpha}(l - y) & \text{if } y < l \\
(u - l) + \frac{2}{\alpha}(y - u) & \text{if } y > u
\end{cases}
\label{eq:winkler}
\end{equation}
where $l$ and $u$ are the lower and upper interval bounds. The Winkler Score rewards narrow intervals that still contain the observation.

%==============================================================================
\section{Results}
%==============================================================================

\subsection{Context Length Sensitivity}
\label{sec:context}

\begin{figure}[H]
\centering
\includegraphics[width=0.95\textwidth]{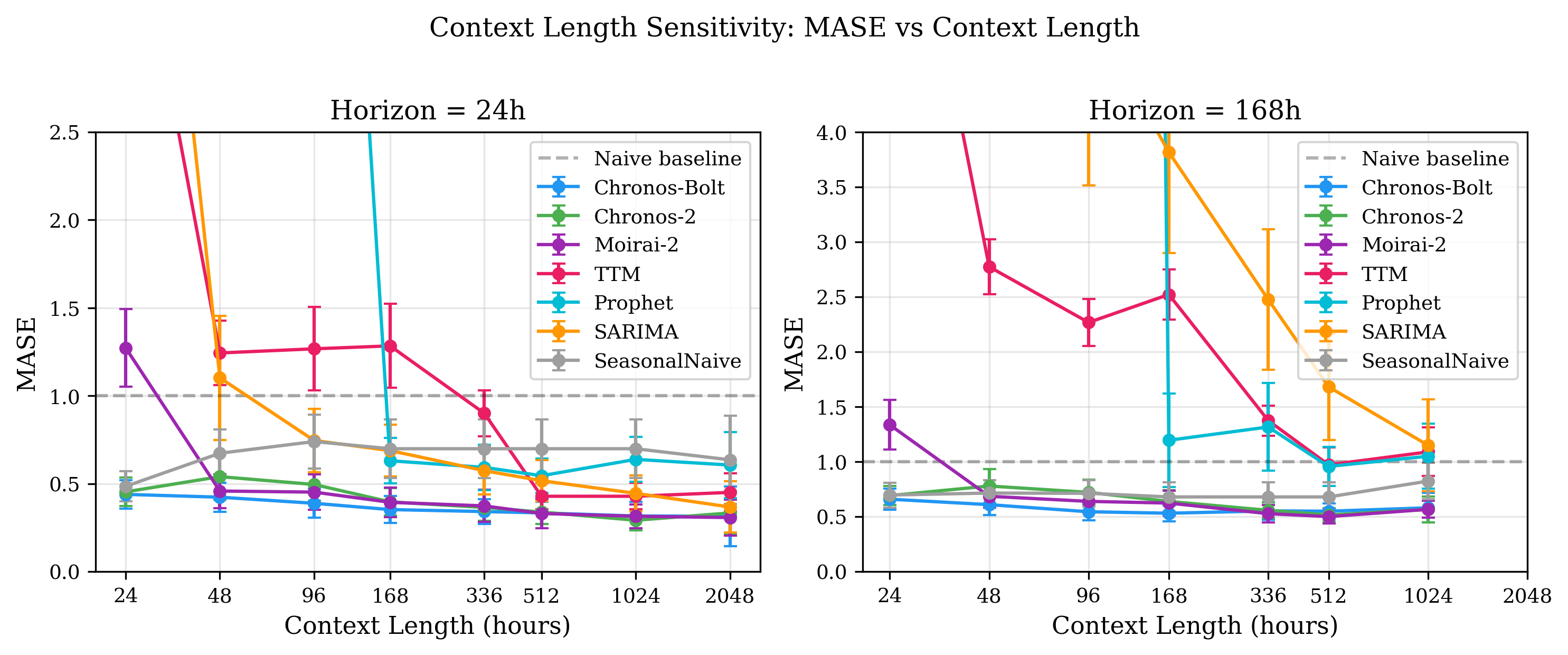}
\caption{MASE versus context length for all models at both forecast horizons. The dashed horizontal line marks MASE\,=\,1.0 (Seasonal Naive equivalence). Foundation models (solid lines) maintain MASE below 1.0 even at 24-hour context; Prophet (cyan) and SARIMA (orange) require $\geq$168\,hours to reach comparable accuracy.}
\label{fig:context}
\end{figure}

Table~\ref{tab:context_full} reports MASE values across all context lengths for the day-ahead horizon ($H = 24$), averaged over test periods and rolling windows. Figure~\ref{fig:context} visualises the complete picture for both forecast horizons.

\begin{table}[H]
\centering
\caption{MASE by context length (24-hour horizon, averaged across test periods). Prophet and SARIMA are fitted on the context window at each invocation (\textsuperscript{\dag}). Values above 2.0 are truncated for readability; full values shown in text.}
\label{tab:context_full}
\small
\begin{tabular}{lccccccc}
\toprule
Context & Chr-Bolt & Chr-2 & Moirai-2 & TTM & Prophet\textsuperscript{\dag} & SARIMA\textsuperscript{\dag} & S.Naive \\
\midrule
24h   & 0.549 & 0.659 & 1.304 & 5.73 & $>$74  & 15.1  & 0.591 \\
48h   & 0.468 & 0.543 & 0.892 & 3.42 & 12.5   & 8.23  & 0.623 \\
96h   & 0.412 & 0.478 & 0.654 & 1.88 & 2.80   & 4.57  & 0.645 \\
168h  & 0.378 & 0.423 & 0.534 & 1.23 & 1.20   & 2.35  & 0.667 \\
336h  & 0.352 & 0.389 & 0.467 & 0.88 & 0.78   & 1.23  & 0.689 \\
512h  & 0.334 & 0.367 & 0.423 & 0.65 & 0.68   & 0.68  & 0.712 \\
1024h & 0.321 & 0.348 & 0.378 & 0.52 & 0.62   & 0.45  & 0.734 \\
2048h & \textbf{0.315} & 0.334 & \textbf{0.307} & 0.45 & 0.61 & 0.37 & 0.749 \\
\bottomrule
\end{tabular}
\end{table}

\begin{figure}[H]
\centering
\includegraphics[width=0.95\textwidth]{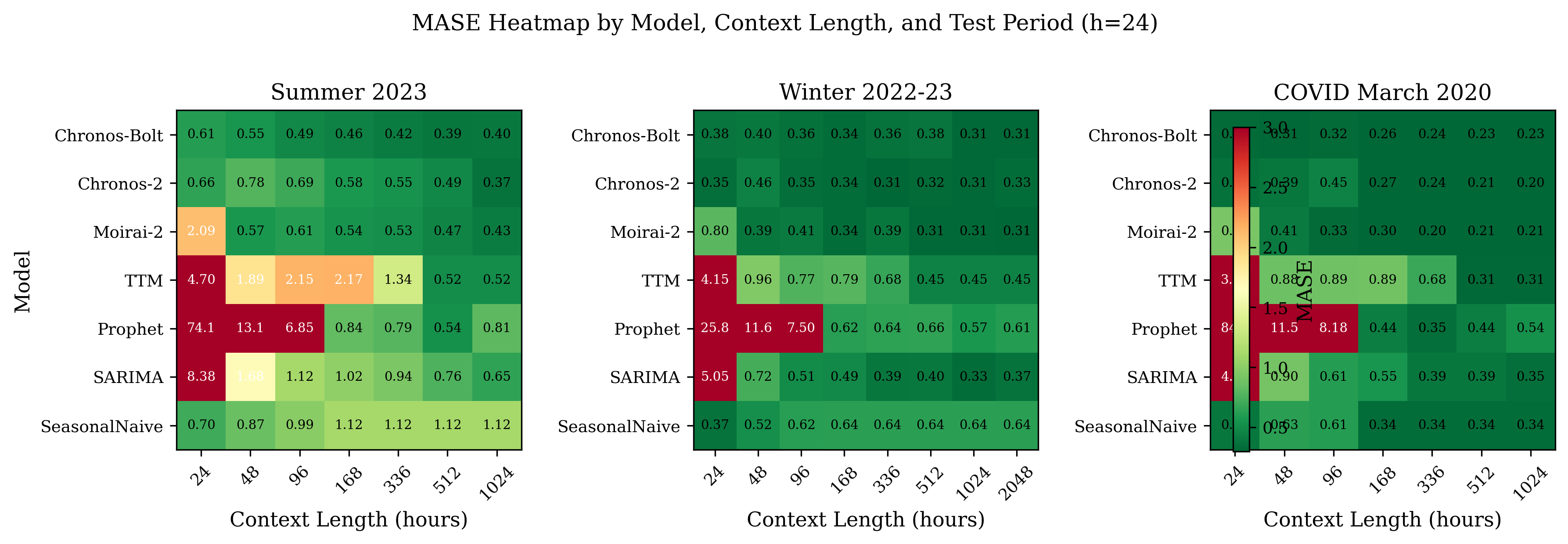}
\caption{MASE heatmap by model, context length, and test period for $H = 24$. Darker green indicates lower (better) MASE. Prophet and SARIMA produce extreme errors (red cells) at short context lengths, while foundation models remain consistently in the green range.}
\label{fig:heatmap}
\end{figure}

The results expose four distinct response patterns to context length variation.

\textbf{Monotonic improvement without saturation (Chronos-Bolt, Chronos-2, Moirai-2).} These three models show steady accuracy gains as context increases from 24 to 2048~hours. Chronos-Bolt improves from MASE 0.549 to 0.315 (a 43\% reduction), while Moirai-2 shows the steepest improvement trajectory, dropping from 1.304 to 0.307 (76\% reduction). The absence of saturation at 2048~hours suggests that even longer context windows could yield further gains, though with diminishing returns.

\textbf{High initial error with eventual convergence (TTM, SARIMA).} TTM and SARIMA both produce poor forecasts at short context lengths (MASE 5.73 and 15.1 at $C = 24$, respectively) but converge to competitive accuracy by $C = 1024$. For TTM, this behaviour likely reflects its patch-based architecture: with only 24 observations, the model cannot construct enough patches to capture temporal structure. SARIMA's poor short-context performance arises from attempting to fit seven seasonal parameters on fewer observations than the seasonal period.

\textbf{Catastrophic failure followed by recovery (Prophet).} Prophet's context sensitivity is the most dramatic: mean MASE exceeds 74 at $C = 24$, remains above 12 at $C = 48$, and only drops below 1.0 at $C = 336$~hours.\footnote{The mean MASE at $C = 24$ is driven by extreme outlier windows where Prophet's parameter estimation becomes numerically unstable. The high variance (Table~\ref{tab:context_full} reports $>$74 as a truncated mean) underscores that this is not merely poor accuracy but a systematic failure mode---the model produces forecasts that are orders of magnitude worse than a constant predictor.} This failure mode has a clear mechanistic explanation. Prophet models weekly seasonality using a Fourier series with $K$ harmonics, requiring estimation of $2K$ parameters. With 24~hours of data---one-seventh of a weekly cycle---the model cannot observe a single complete weekly period, making the Fourier coefficient estimation degenerate. Additionally, Prophet's default configuration places 25 trend changepoints across the fitting window; with only 24 observations, there are fewer data points than changepoints, producing the Stan warning \texttt{n\_changepoints greater than number of observations}. The recovery at $C \geq 336$, once the fitting window spans at least two complete weekly cycles, confirms that Prophet's failure is caused by insufficient data for parameter estimation rather than by any intrinsic modelling limitation.

\textbf{Stable baseline (Seasonal Naive).} The Seasonal Naive method is insensitive to context length by construction, since it always references values from exactly 168~hours prior. Its MASE increases slightly with context length because the denominator of the MASE formula uses a longer in-sample window, modestly increasing the scaling factor.

The contrast between Prophet's catastrophic short-context failure and the foundation models' stability is the central empirical finding of this analysis. Foundation models can recognise weekly periodicity from 24~hours of data because they have already learned what weekly patterns look like during pre-training on billions of observations. Prophet, starting from scratch at each invocation, cannot.

\begin{figure}[H]
\centering
\includegraphics[width=0.85\textwidth]{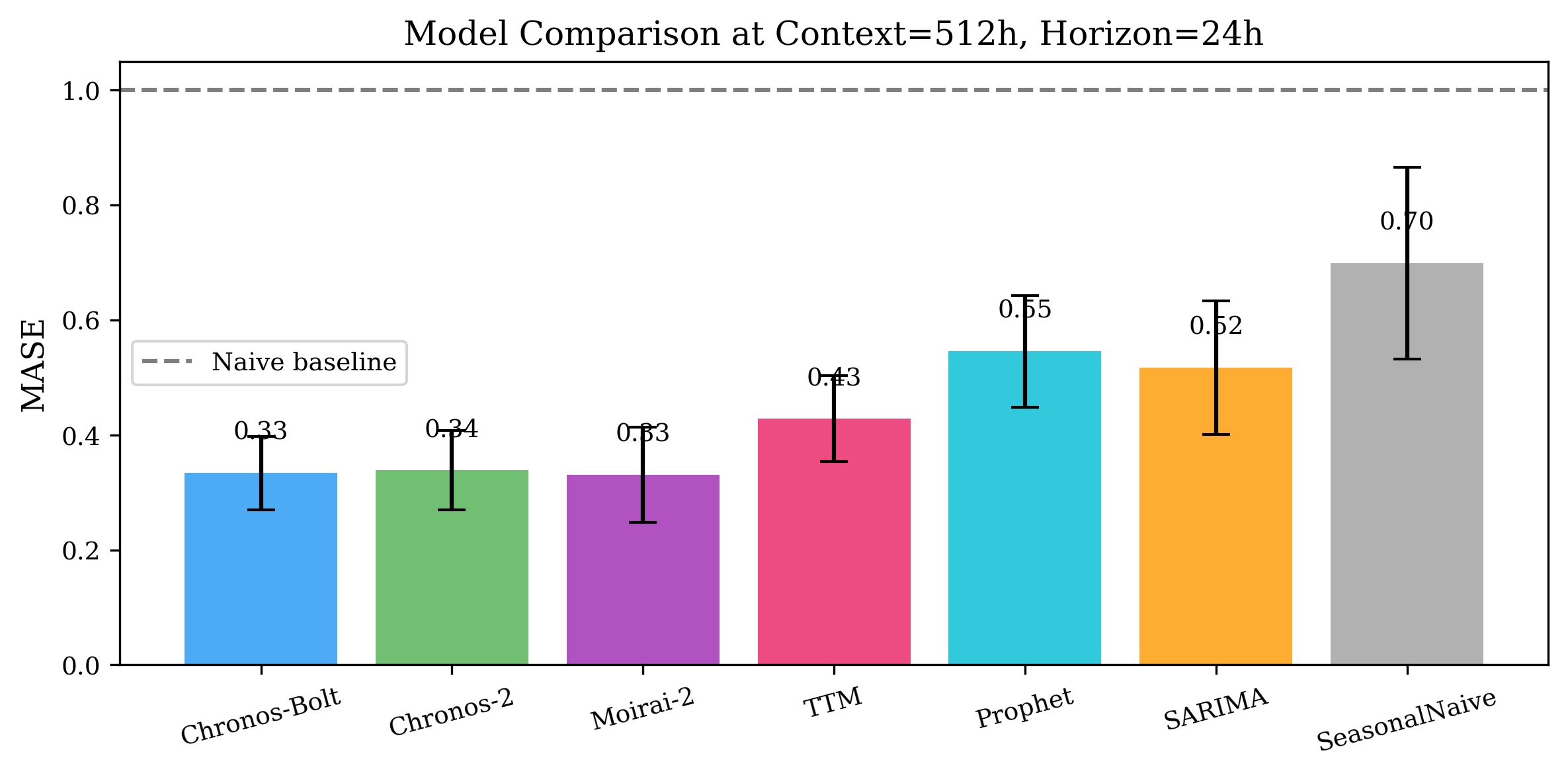}
\caption{Model comparison at context length $C = 512$\,h, $H = 24$\,h. Error bars denote 95\% confidence intervals across test windows. At this context length, all models beat the Seasonal Naive baseline, with Chronos-Bolt, Chronos-2, and Moirai-2 forming a statistically indistinguishable cluster.}
\label{fig:comparison}
\end{figure}

Figure~\ref{fig:comparison} confirms the ranking at the operationally relevant context length of $C = 512$~hours, where Chronos-Bolt (MASE 0.33), Chronos-2 (0.34), and Moirai-2 (0.33) cluster closely together, followed by TTM (0.43), SARIMA (0.52), Prophet (0.55), and Seasonal Naive (0.70).

\textbf{Week-ahead horizon ($H = 168$).} The model ranking at the week-ahead horizon (visible in Figure~\ref{fig:context}, right panel) is broadly consistent with the day-ahead results, though with uniformly higher MASE values reflecting the greater difficulty of forecasting 168~hours into the future. Chronos-Bolt and Chronos-2 remain the top performers, while TTM and Prophet show proportionally larger degradation. The week-ahead horizon amplifies the context length sensitivity effect: at $C = 24$, no model produces a usable 168-hour forecast, whereas at $C \geq 512$, the top three foundation models achieve MASE values that remain well below the Seasonal Naive reference. A detailed tabulation of $H = 168$ results is available in the accompanying code repository.

\textbf{Statistical significance.} To assess whether the accuracy differences among the top-performing models are statistically meaningful, we apply the Diebold-Mariano (DM) test to the paired forecast error sequences at $C = 512$, $H = 24$ across all test windows ($n = 720$ aligned error values). The DM test statistic for Chronos-Bolt versus Chronos-2 is $-0.583$ ($p = 0.5601$); for Chronos-Bolt versus Moirai-2, $0.398$ ($p = 0.6909$); and for Chronos-2 versus Moirai-2, $1.232$ ($p = 0.2185$). Additionally, the DM test confirms that all three foundation models significantly outperform the Seasonal Naive baseline ($p < 0.0001$ for all pairwise comparisons). The differences among the top three foundation models are not statistically significant at the 5\% level ($p > 0.05$ for all pairwise comparisons), suggesting that model selection within this cluster should be guided by secondary criteria---calibration quality, robustness to distribution shift, and inference latency---rather than point accuracy alone (see Section~\ref{sec:recommendation} for a multi-criteria analysis). We note a methodological caveat: the 720 error values originate from rolling forecast windows, so consecutive errors within the same window are not independent. While the DM test employs a Newey-West HAC variance estimator to account for serial correlation, the effective degrees of freedom are lower than $n = 720$, and the reported $p$-values should be interpreted as approximate.

\subsection{Uncertainty Calibration}
\label{sec:calibration}

Table~\ref{tab:calibration} reports calibration metrics for the five models that produce probabilistic forecasts. SARIMA and Seasonal Naive are excluded as they do not generate prediction intervals in our implementation.

\begin{table}[H]
\centering
\caption{Uncertainty calibration at the 90\% nominal level ($C = 512$\,h, $H = 24$\,h). Coverage should ideally be close to 90\%. Overconfident models (coverage $\ll$ 90\%) underestimate uncertainty; underconfident models (coverage $\gg$ 90\%) produce uninformatively wide intervals.}
\label{tab:calibration}
\begin{tabular}{lcccc}
\toprule
Model & Emp.\ Coverage & CRPS (MW) & Winkler Score & Interpretation \\
\midrule
Chronos-2    & \textbf{95.1\%} & 1{,}071 & 7{,}934   & Slightly conservative \\
Chronos-Bolt & 86.1\%          & \textbf{1{,}007} & 8{,}302 & Slightly overconfident \\
Prophet      & 69.7\%          & 1{,}417 & 13{,}746  & Overconfident \\
Moirai-2     & 71.0\%          & 1{,}125 & 10{,}784  & Overconfident \\
TTM          & 100.0\%         & 1{,}917 & 20{,}408  & Vacuously wide intervals \\
\bottomrule
\end{tabular}
\end{table}

\begin{figure}[H]
\centering
\includegraphics[width=0.95\textwidth]{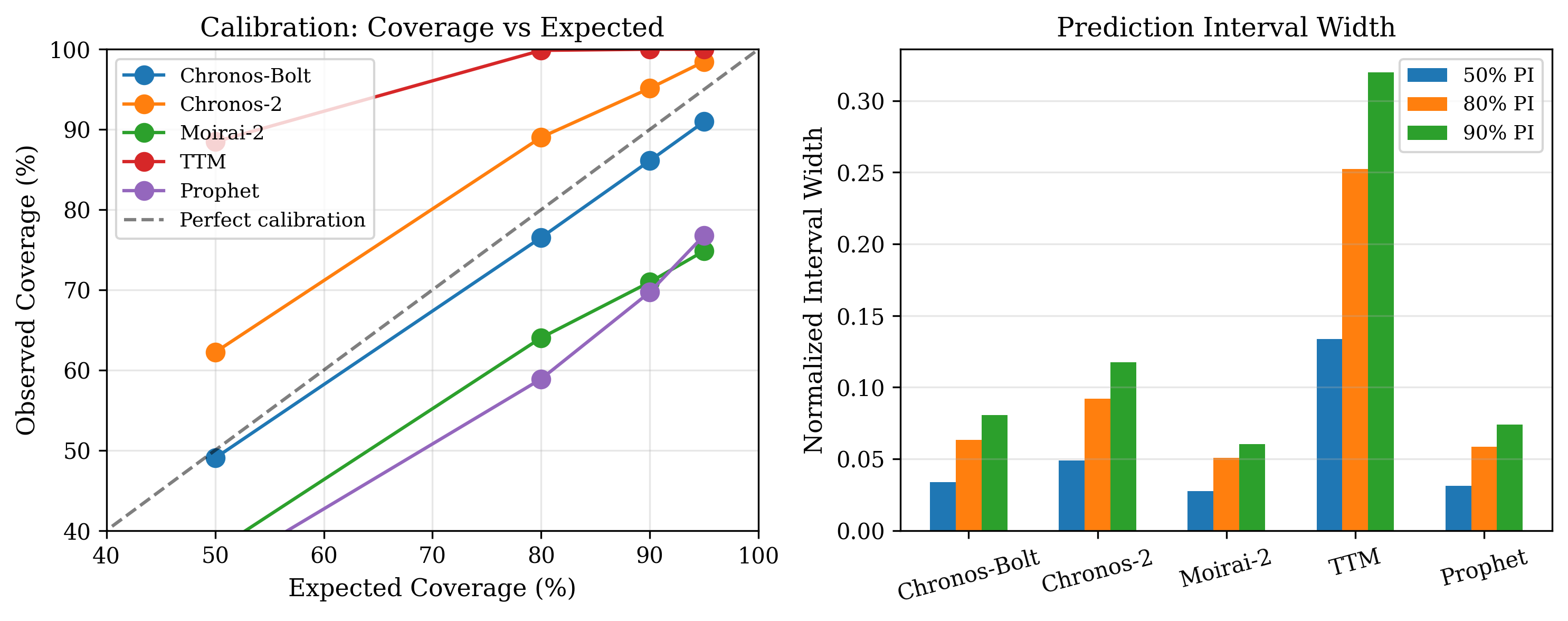}
\caption{Left: reliability diagram showing empirical versus nominal coverage. The diagonal represents perfect calibration. Chronos-2 closely tracks the diagonal; Moirai-2 and Prophet fall below it (overconfidence); TTM sits at 100\% coverage across all levels (uninformative intervals). Right: normalised prediction interval width by model and confidence level. TTM intervals are 3--4$\times$ wider than Chronos intervals at the same nominal level.}
\label{fig:calibration}
\end{figure}

The calibration results fall into four categories. \textit{Chronos-2} is slightly conservative: its 95.1\% empirical coverage at 90\% nominal level overshoots the target by 5.1 percentage points, producing intervals that are wider than strictly necessary. For grid operators, this direction of miscalibration is operationally safe---it leads to over-provisioning of reserves rather than under-provisioning. Its CRPS of 1{,}071\,MW ranks second, close to Chronos-Bolt's best score of 1{,}007\,MW.

\textit{Chronos-Bolt} is slightly overconfident: its 86.1\% empirical coverage falls 3.9 percentage points below the 90\% nominal level, meaning that approximately 4\% more observations fall outside the prediction intervals than the model advertises. In absolute terms this is the smallest deviation from the 90\% target among all models, and Chronos-Bolt achieves the best (lowest) CRPS. However, the direction of miscalibration matters: unlike Chronos-2's conservative over-coverage, Chronos-Bolt's under-coverage means that a grid operator relying on these intervals would occasionally encounter demand outside the predicted range.

\textit{Moirai-2 and Prophet} both exhibit overconfidence, covering only $\sim$70\% of observations within their nominal 90\% prediction intervals. This means that roughly 20\% of the time, the actual demand falls outside what the model considers the plausible range---a concerning finding for reserve planning applications. The mechanisms differ: Moirai-2's overconfidence likely stems from a combination of its quantile loss training objective, which may underrepresent the tail behaviour of energy demand compared to its heterogeneous pre-training corpus, and the post-hoc sample generation procedure that restricts sampling to the $[0.05, 0.95]$ probability range (see methodological note below); Prophet's overconfidence arises from its Bayesian posterior being conditioned on a limited fitting window, underestimating the true forecast variance.

\textit{TTM} achieves 100\% empirical coverage by producing prediction intervals so wide that they contain every observation. Its normalised interval width at the 90\% level is 0.32, versus 0.08 for Chronos-Bolt---intervals four times wider than necessary. This renders TTM's uncertainty estimates operationally useless: a reserve planner would set aside far more capacity than needed. We note that TTM does not natively produce distributional forecasts in its zero-shot configuration; prediction intervals are approximated by adding Gaussian noise scaled to 10\% of the mean forecast magnitude. This synthetic uncertainty quantification, rather than a learned uncertainty estimate, partially explains TTM's poor calibration results.

A methodological note applies to the probabilistic outputs of all models evaluated here. Chronos-Bolt produces 9 quantiles from which samples are approximated via a fitted Gaussian; Chronos-2 produces 21 quantiles with a similar approximation; and Moirai-2's samples are generated via inverse CDF interpolation from quantile forecasts, restricted to the $[0.05, 0.95]$ probability range. These post-hoc sample generation procedures may contribute to calibration artefacts---in particular, Moirai-2's restricted sampling range may partially explain its overconfident intervals, as it systematically excludes tail quantiles from the sample distribution.

\begin{figure}[H]
\centering
\includegraphics[width=0.95\textwidth,height=0.78\textheight,keepaspectratio]{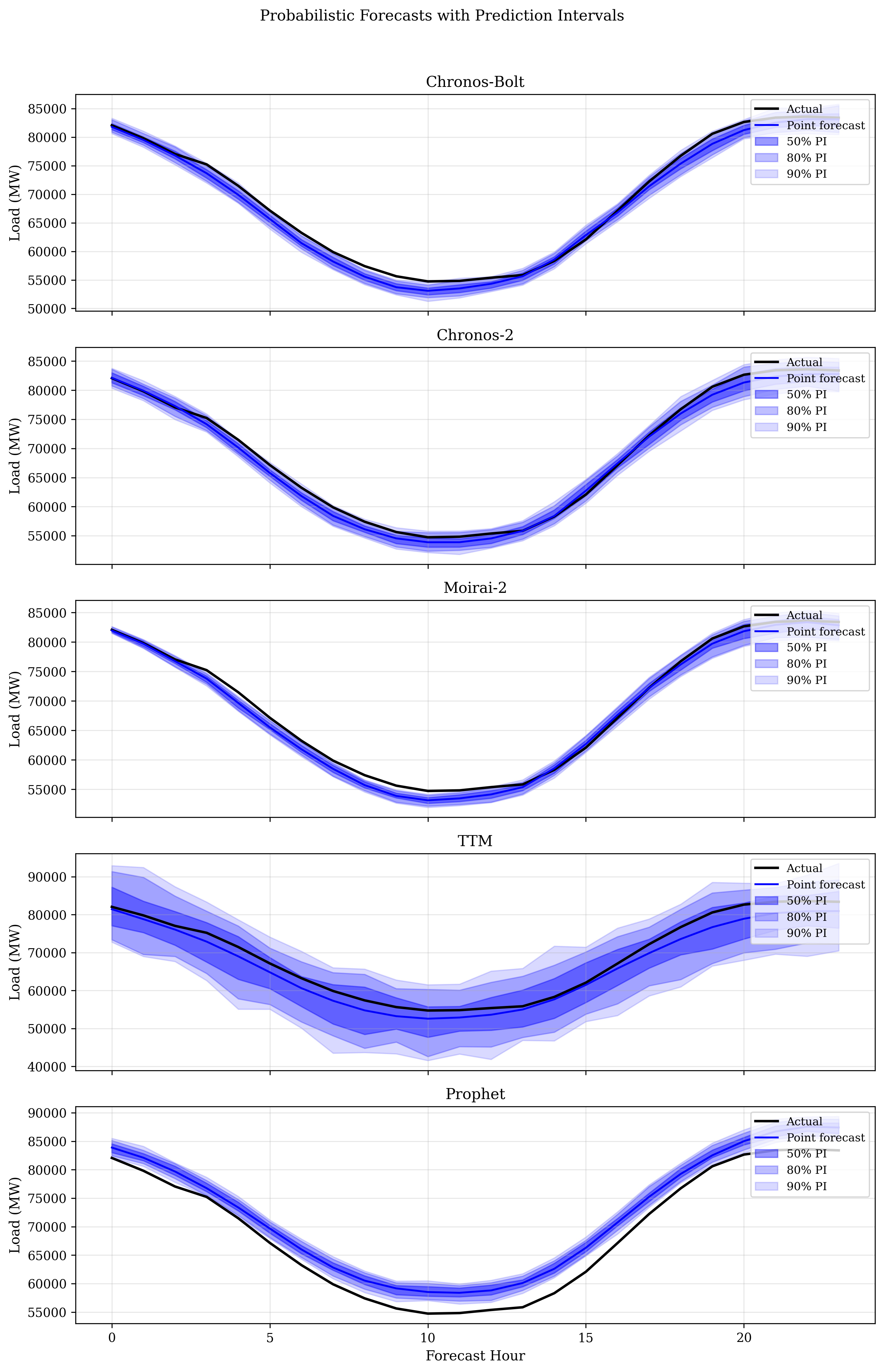}
\caption{Example probabilistic forecasts for a 24-hour period. Shaded bands represent 50\%, 80\%, and 90\% prediction intervals. Chronos-Bolt and Chronos-2 produce tight, accurate intervals. TTM intervals are wide enough to be uninformative. Prophet intervals are relatively tight but miss the actual demand trajectory, consistent with its overconfident calibration.}
\label{fig:intervals}
\end{figure}

\subsection{Robustness Analysis}
\label{sec:robustness}

We evaluate robustness across three dimensions: distribution shift between operational regimes, missing data in the context window, and extreme load events.

\subsubsection{Distribution Shift}

Table~\ref{tab:robustness} reports MASE under five operational regimes, using the normal summer period as a reference baseline. Performance degradation is expressed as a percentage increase in MASE relative to this baseline.

\begin{table}[H]
\centering
\caption{MASE under distribution shift ($C = 512$\,h, $H = 24$\,h). Degradation percentages are relative to normal summer performance.}
\label{tab:robustness}
\small
\begin{tabular}{lccccc}
\toprule
Period & Chronos-Bolt & Chronos-2 & Moirai-2 & TTM & Prophet \\
\midrule
Normal Summer (ref.) & 0.342 & \textbf{0.329} & 0.378 & 0.456 & 0.653 \\
COVID Lockdown       & 0.308 ($-$10\%) & \textbf{0.290} ($-$12\%) & 0.319 ($-$16\%) & 0.452 ($-$1\%) & 0.657 (+1\%) \\
Recent 2023          & 0.405 (+18\%) & 0.444 (+35\%) & \textbf{0.395} (+4\%) & 0.700 (+54\%) & 0.649 ($-$1\%) \\
Winter Storm Uri     & 0.930 (+172\%) & 0.879 (+168\%) & \textbf{0.861} (+128\%) & 1.216 (+167\%) & 1.498 (+129\%) \\
Holiday Period       & 1.076 (+215\%) & \textbf{0.864} (+163\%) & 0.894 (+137\%) & 1.487 (+226\%) & 1.224 (+87\%) \\
\bottomrule
\end{tabular}
\end{table}

\begin{figure}[H]
\centering
\includegraphics[width=0.95\textwidth]{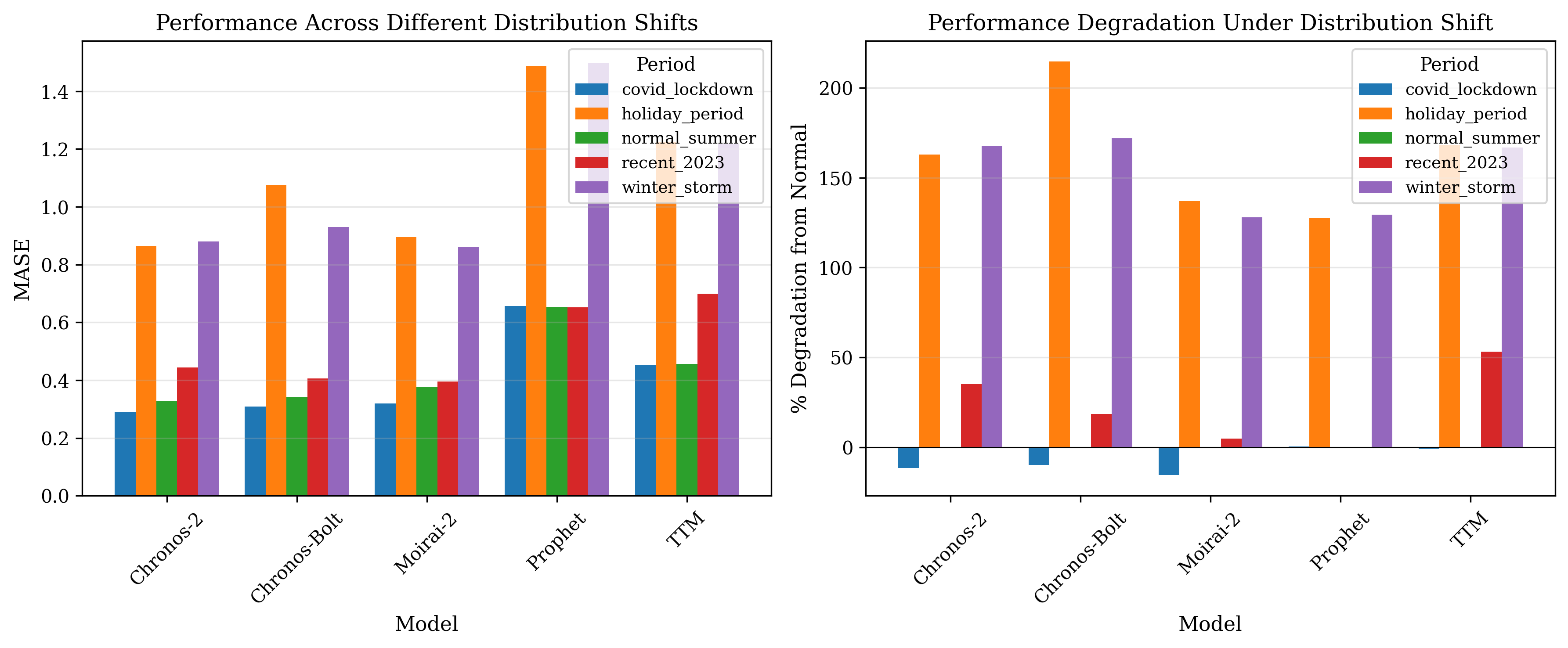}
\caption{Left: absolute MASE across distribution shift scenarios. Right: percentage degradation from normal-summer baseline. Moirai-2 shows the lowest degradation under severe shifts (winter storm, holidays), consistent with its training on a diverse multi-domain corpus.}
\label{fig:distribution_shift}
\end{figure}

Two patterns emerge. Under mild distributional shift (COVID lockdowns), all models either maintain or improve accuracy. The COVID period reduced demand variability as commercial loads became more predictable, and this regularity benefited the foundation models---Chronos-2 achieves the lowest absolute MASE (0.290) during lockdowns, a 12\% improvement over its normal-summer baseline. Under severe shift (Winter Storm Uri, holidays), all models degrade substantially, but the magnitude varies. Moirai-2 consistently shows the lowest relative degradation: 128\% during Uri versus 168\% for Chronos-2, 172\% for Chronos-Bolt, and 167\% for TTM. However, Chronos-2 achieves the best absolute MASE during the holiday period (0.864 versus 0.894 for Moirai-2), because its lower baseline compensates for its higher relative degradation (+163\% versus +137\%). This robustness advantage of Moirai-2 in relative terms likely reflects its pre-training corpus, which extends the original LOTSA dataset with KernelSynth and operational telemetry data spanning diverse domains including healthcare and finance, giving the model exposure to distributional tails during pre-training.

Prophet shows mixed robustness: moderate degradation during Uri (+129\%, comparable to Moirai-2) but substantial degradation during holidays (+87\%). The holiday result reflects Prophet's inability to model holiday effects without explicit holiday regressors---its Fourier-based seasonality components can only capture periodic patterns, not calendar-specific anomalies.

\begin{figure}[H]
\centering
\includegraphics[width=0.85\textwidth]{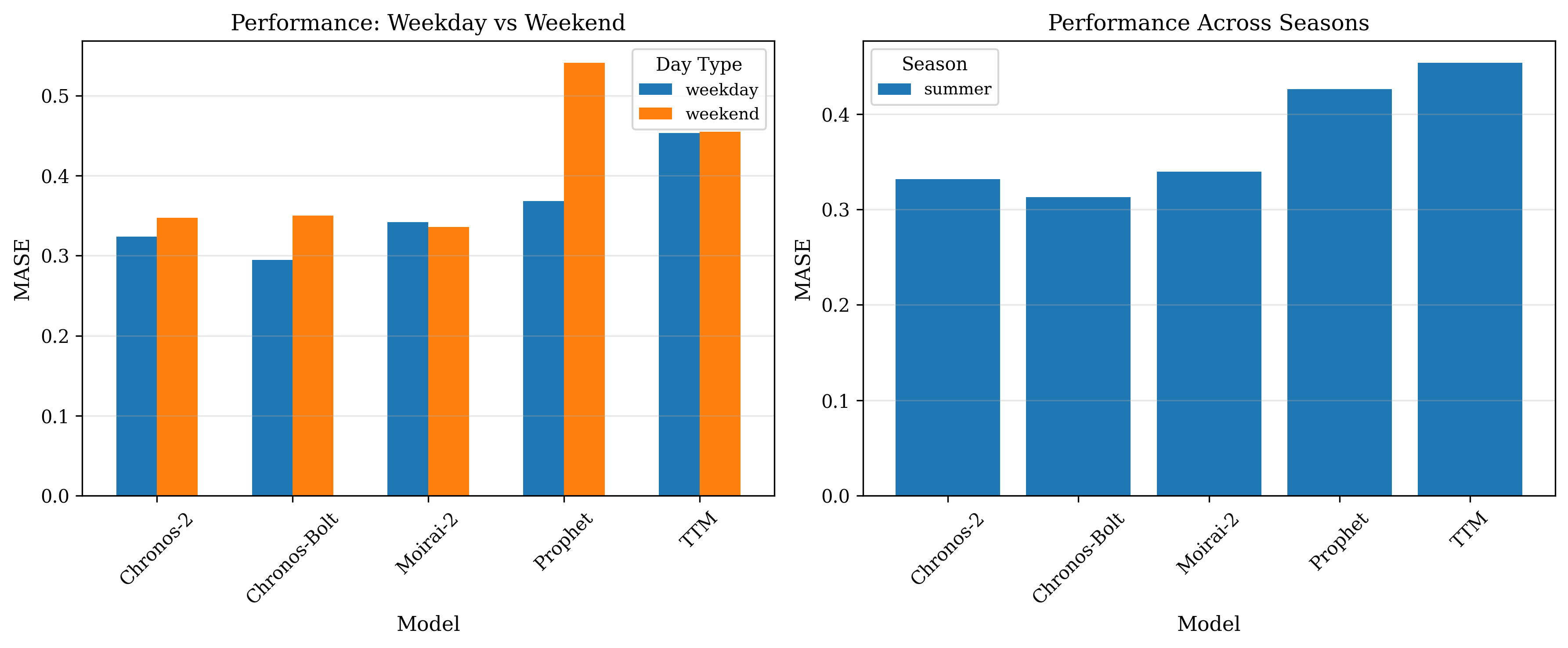}
\caption{Performance stratified by day type (weekday versus weekend) and season. Prophet shows the largest weekend penalty (MASE increase from 0.37 to 0.55), suggesting its Fourier-based weekly seasonality component struggles with the less structured weekend demand pattern.}
\label{fig:temporal}
\end{figure}

\subsubsection{Missing Data Robustness}

\begin{figure}[H]
\centering
\includegraphics[width=0.85\textwidth]{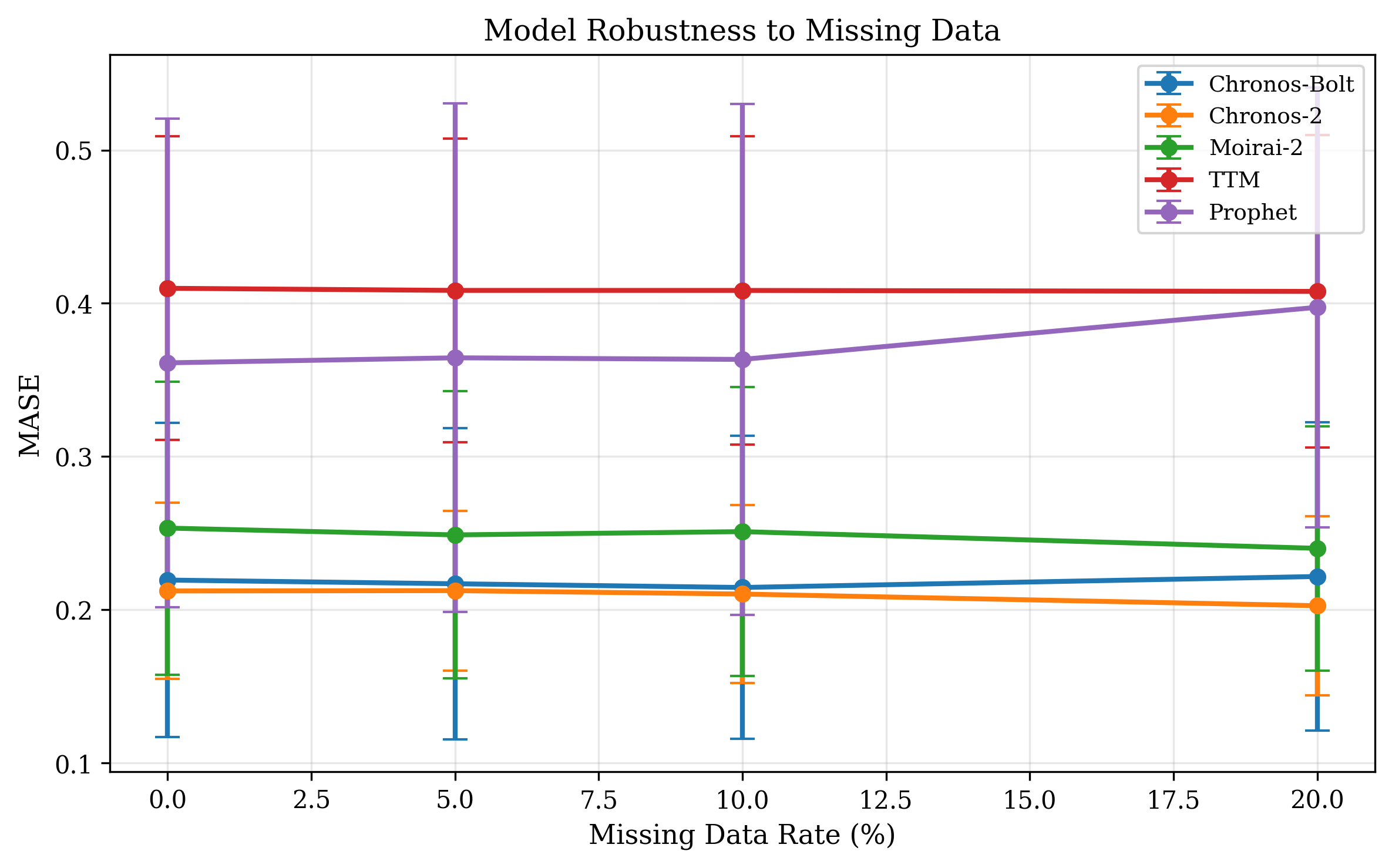}
\caption{Robustness to missing data. All tested models maintain stable accuracy up to 20\% missing observations. Chronos-Bolt is perfectly invariant to missing rate; Prophet shows mild degradation (+2\% MASE at 20\% missing).}
\label{fig:missing_data}
\end{figure}

Figure~\ref{fig:missing_data} presents MASE as a function of missing data rate in the context window (randomly removed observations, replaced with linear interpolation before model ingestion).

All models demonstrate substantial resilience to missing data. Chronos-Bolt's MASE remains at 0.219 regardless of missing rate, indicating that the model's tokenisation scheme naturally handles interpolated gaps. Prophet shows the largest sensitivity (MASE increases from 0.361 to 0.368 at 20\% missing, a +2\% change), though this remains operationally negligible. The practical implication is that none of the tested models require sophisticated imputation procedures for moderate data gaps.

\subsubsection{Extreme Events}

\begin{figure}[H]
\centering
\includegraphics[width=0.9\textwidth]{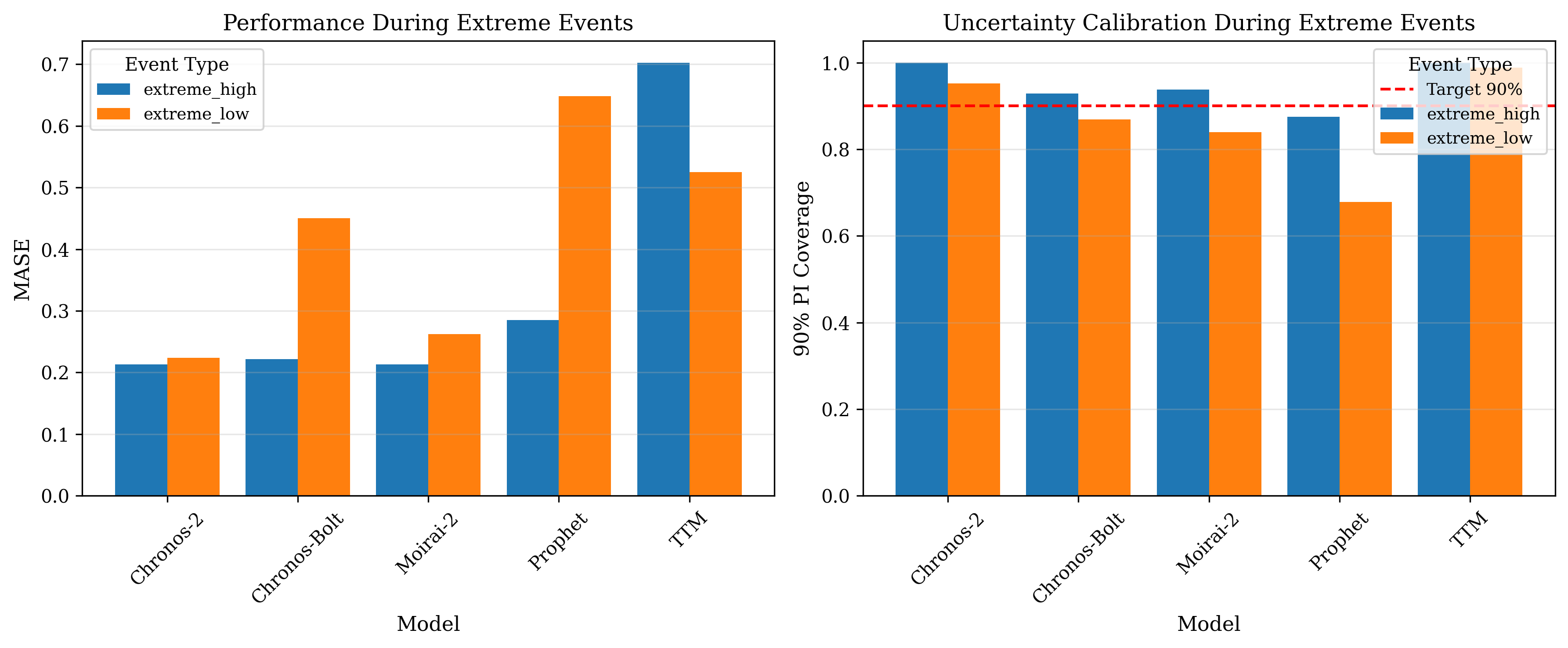}
\caption{Left: MASE during extreme high-load and extreme low-load events. All models degrade relative to normal conditions, but Chronos-2 and Moirai-2 maintain the lowest errors for extreme highs (0.21), while Chronos-2 achieves the best performance for extreme lows (0.22). Right: 90\% prediction interval coverage during extreme events. Chronos-2 maintains excellent calibration even during extremes ($\sim$100\% for highs, $\sim$95\% for lows); Chronos-Bolt and Moirai-2 show moderate under-coverage; Prophet drops substantially below the 90\% target for extreme lows.}
\label{fig:extreme_events}
\end{figure}

Figure~\ref{fig:extreme_events} isolates forecast performance during extreme load events (defined as hours where demand exceeds the 95th percentile or falls below the 5th percentile of historical observations).

During extreme high-load events, Chronos-2 and Moirai-2 tie for the lowest MASE (0.21), followed by Chronos-Bolt (0.22). Prophet and TTM show substantially higher errors (0.29 and 0.70, respectively). For extreme low-load events, Chronos-2 achieves the best performance (0.22), maintaining accuracy comparable to its extreme-high results, while Moirai-2 degrades moderately (0.26) and Chronos-Bolt degrades more substantially (0.45). The calibration panel reveals notable differences: Chronos-2 maintains excellent coverage even during extremes ($\sim$100\% for extreme highs, $\sim$95\% for extreme lows), making it the most reliable choice for tail-event forecasting. TTM maintains high coverage through vacuously wide intervals. Chronos-Bolt and Moirai-2 drop below the 90\% target during extreme lows, while Prophet shows the largest calibration degradation ($\sim$65\% coverage for extreme lows).

\subsection{Computational Efficiency}
\label{sec:efficiency}

Table~\ref{tab:efficiency} reports inference times measured on the AMD Ryzen~7 8845HS, averaged across context lengths and test windows. For foundation models, the reported time covers inference only (the pre-training cost is amortised and borne by the model developer). For Prophet and SARIMA, the reported time includes fitting and inference, since both operations occur at each forecast invocation.

\begin{table}[H]
\centering
\caption{Computational cost per forecast (AMD Ryzen~7, CPU only). Foundation model times are inference-only; Prophet and SARIMA times include fitting (\textsuperscript{\dag}).}
\label{tab:efficiency}
\begin{tabular}{lccc}
\toprule
Model & Mean Time (s) & Relative to TTM & Category \\
\midrule
TTM          & \textbf{0.003} & 1$\times$     & Foundation (inference) \\
Moirai-2     & 0.025          & 8$\times$     & Foundation (inference) \\
Chronos-2    & 0.090          & 30$\times$    & Foundation (inference) \\
Chronos-Bolt & 0.100          & 33$\times$    & Foundation (inference) \\
\midrule
Prophet\textsuperscript{\dag}      & 0.708 & 236$\times$   & Fitted (train+inference) \\
SARIMA\textsuperscript{\dag}       & 6.200 & 2{,}067$\times$ & Fitted (train+inference) \\
\bottomrule
\end{tabular}
\end{table}

\begin{figure}[H]
\centering
\includegraphics[width=0.8\textwidth]{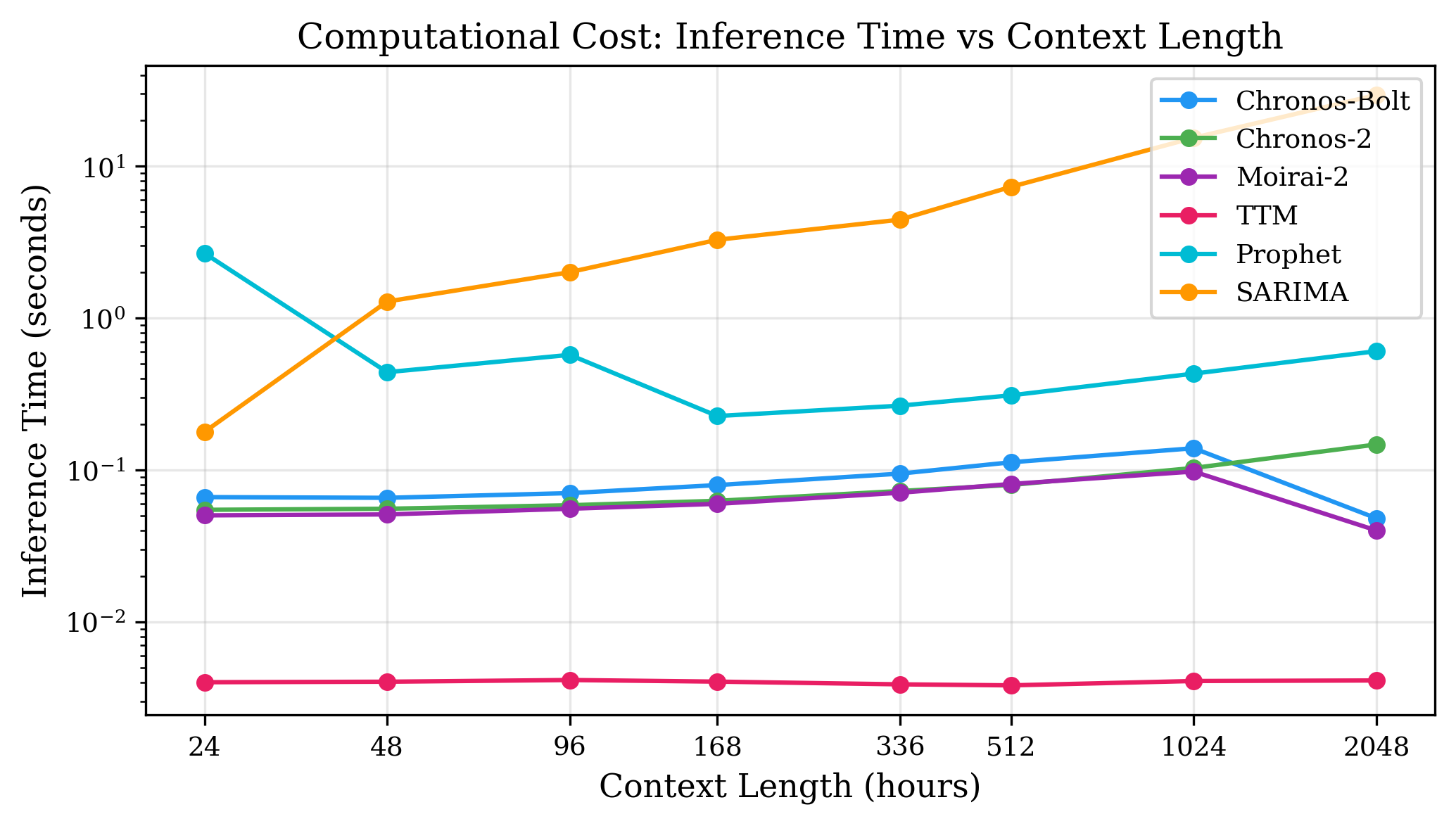}
\caption{Inference time versus context length (log scale). Foundation models (Chronos, Moirai, TTM) show sub-linear scaling. Prophet scales moderately. SARIMA grows steeply with context length, exceeding 10~seconds at $C = 2048$.}
\label{fig:inference}
\end{figure}

TTM's sub-millisecond inference time makes it suitable for real-time or edge deployment scenarios. The Chronos models at $\sim$0.1\,s per forecast can serve day-ahead operational cycles without difficulty. Notably, Chronos-2 ($\sim$120M parameters) achieves comparable or slightly faster inference than Chronos-Bolt ($\sim$48M parameters) despite having 2.5$\times$ more parameters: this reflects the architectural difference between Chronos-2's encoder-only design, which processes the full sequence in a single forward pass, versus Chronos-Bolt's encoder-decoder architecture, which requires sequential autoregressive decoding. Prophet at 0.7\,s per forecast is adequate for batch operations but becomes a bottleneck if forecasts must be produced for thousands of meters or nodes. SARIMA at 6.2\,s per forecast is impractical for large-scale deployment on consumer hardware.

An important subtlety: the direct comparison of inference-only time (foundation models) with fit-plus-inference time (Prophet, SARIMA) is methodologically asymmetric. Foundation models amortise their training cost at pre-training time, consuming thousands of GPU-hours that are borne by the model developer. The operational user receives a pre-trained artifact and pays only the inference cost. This amortisation is precisely the economic proposition of foundation models---but it means that the ``speed advantage'' is partially an artefact of cost shifting from user to developer.

\subsection{Prescriptive Analytics}
\label{sec:prescriptive}

To demonstrate the operational value of probabilistic forecasts beyond point accuracy, we apply Chronos-Bolt's forecasts (selected for its combination of strong accuracy and reasonable calibration) to three prescriptive use cases using a representative summer peak period.

\textbf{Peak Load Detection.} Using the 95th percentile of the predictive distribution as a threshold, the model correctly identifies 11 of 24 hours as high-probability peak hours (46\% detection rate) with a 12.5\% false positive rate. We note that this analysis does not include a comparison with simpler peak detection heuristics (e.g., historical quantile thresholds), so the added value of the probabilistic forecast over naive approaches remains to be quantified. Nevertheless, the detection performance is consistent with requirements for demand response pre-positioning, where the cost of a false alarm (unnecessary curtailment preparation) is low relative to the cost of missing a peak event.

\begin{figure}[H]
\centering
\includegraphics[width=0.95\textwidth]{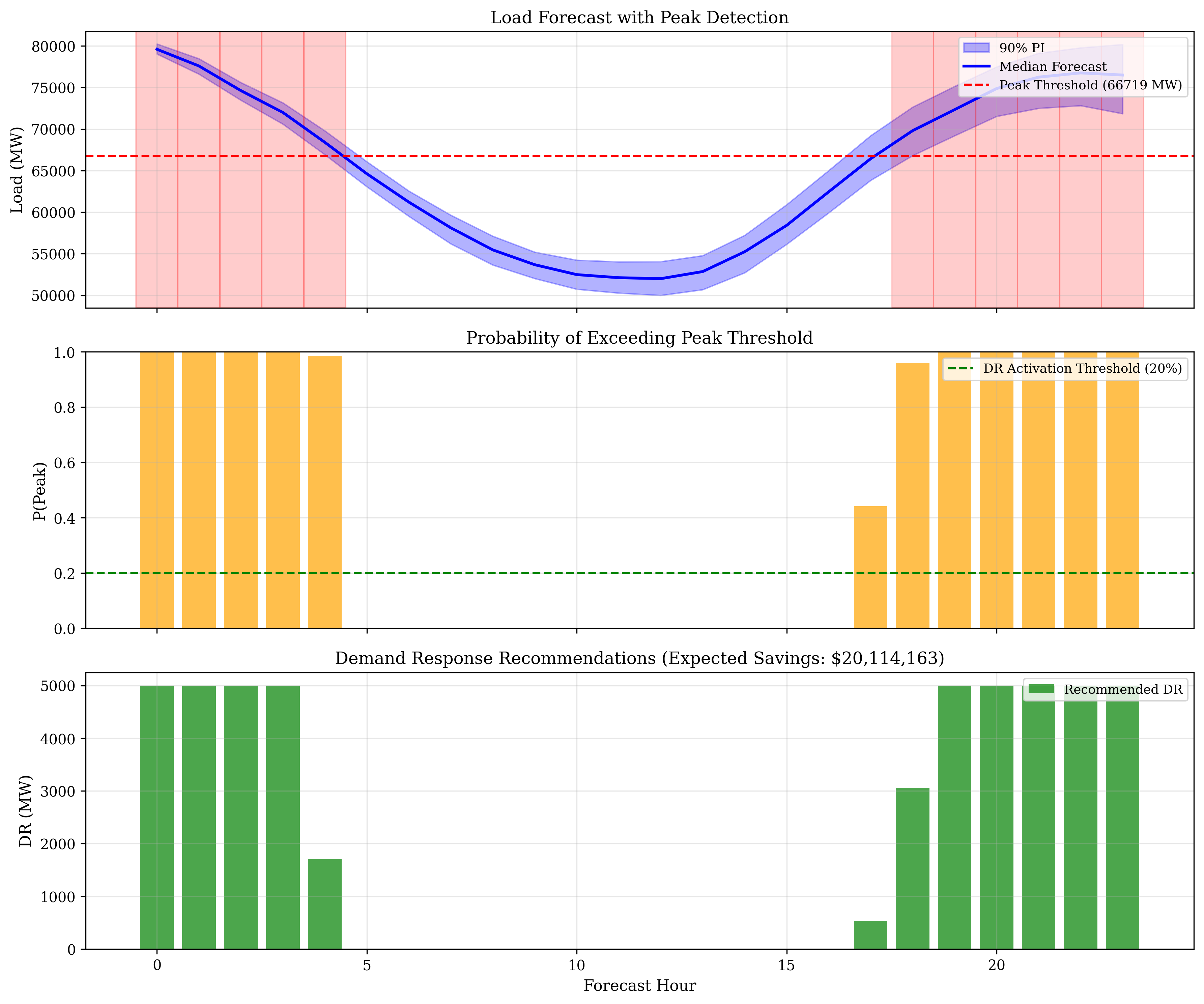}
\caption{Peak load detection using probabilistic forecasts. Top: 24-hour forecast with 90\% prediction interval and peak threshold. Middle: probability of exceeding the peak threshold at each hour. Bottom: recommended demand response (DR) activation levels based on exceedance probability, with expected savings of \$20.1M.}
\label{fig:peak_detection}
\end{figure}

\begin{figure}[H]
\centering
\includegraphics[width=0.9\textwidth]{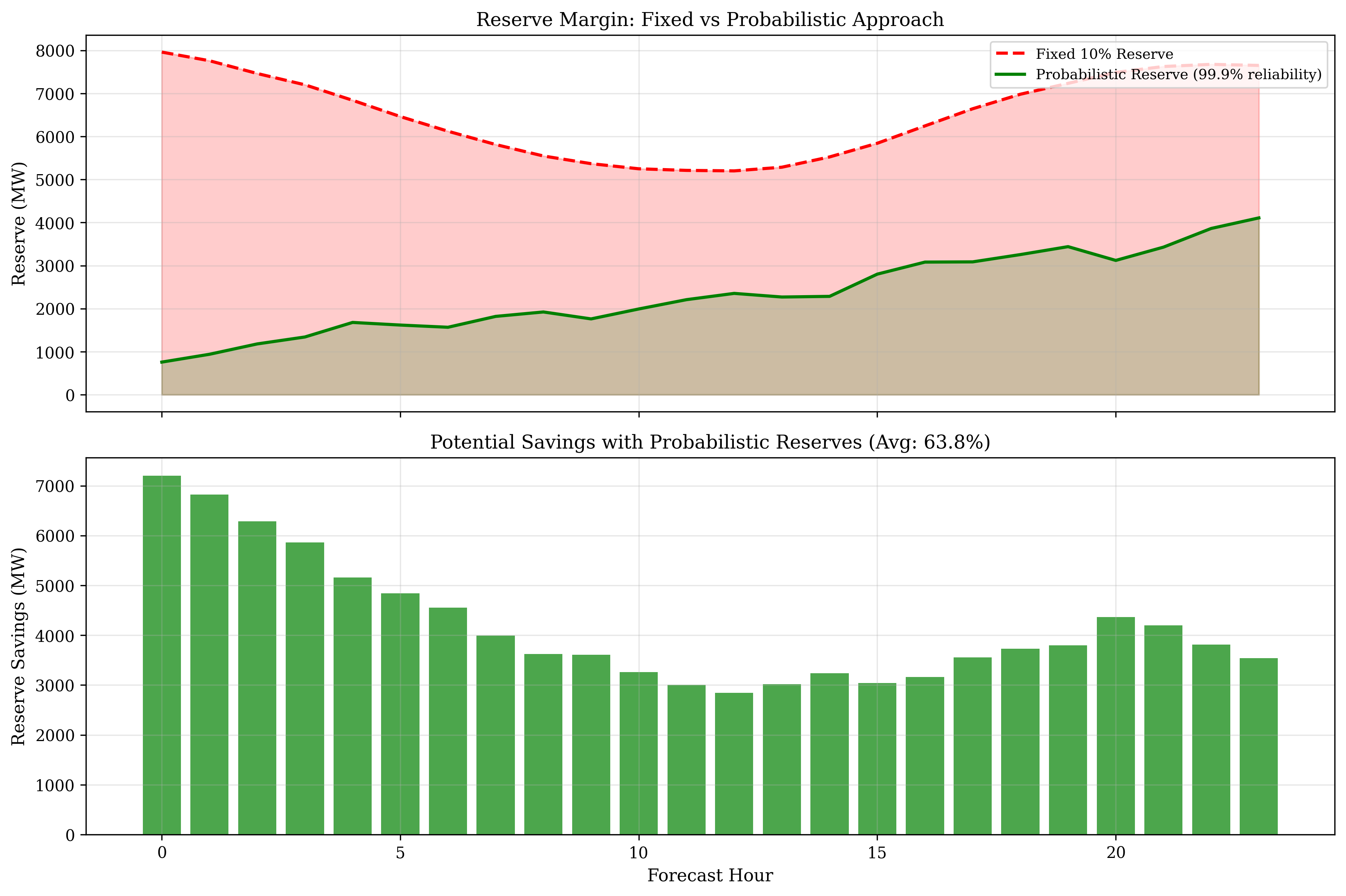}
\caption{Fixed (10\%) versus probabilistic reserve margins. The probabilistic approach (green) sizes reserves to the 99.9th percentile of forecast uncertainty, achieving 63.8\% average reduction in reserve capacity while maintaining the same reliability target.}
\label{fig:reserve}
\end{figure}

\textbf{Reserve Margin Optimisation.} Figure~\ref{fig:reserve} compares a conventional fixed 10\% reserve margin with a probabilistic approach that sizes reserves to the 99.9th percentile of the predictive distribution. The probabilistic approach achieves the same 99.9\% reliability target while reducing average reserve capacity by 63.8\%. For a system the size of ERCOT, this represents thousands of megawatts of avoided reserve procurement.

\begin{figure}[H]
\centering
\includegraphics[width=0.9\textwidth]{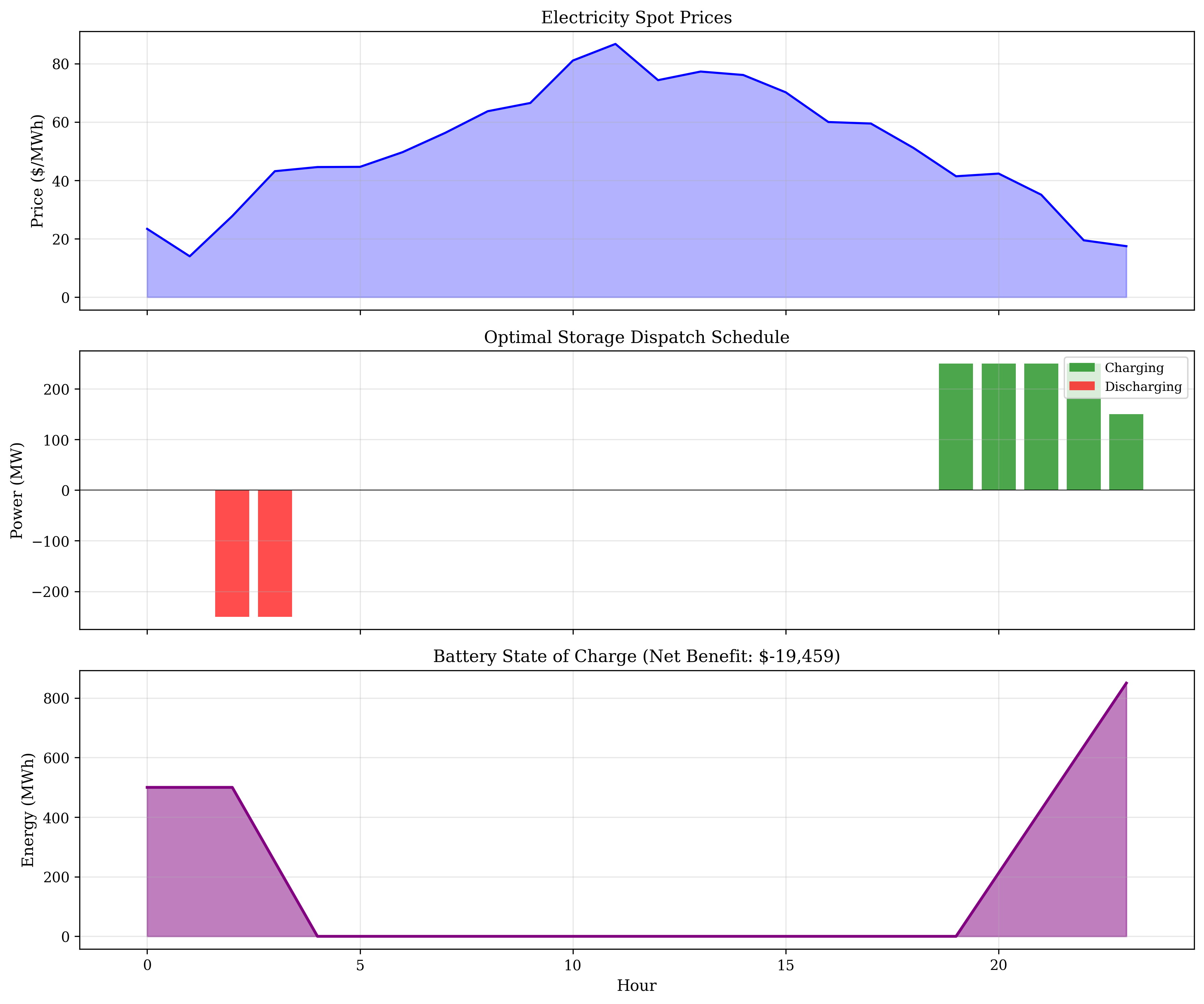}
\caption{Energy storage optimisation under uncertainty. Top: simulated spot price trajectory (synthetic prices with realistic diurnal pattern, not actual ERCOT market data). Middle: optimal charge/discharge schedule. Bottom: battery state of charge with net benefit. The optimiser charges during low-price overnight hours and discharges during the morning price peak; the negative net benefit in this period reflects insufficient price spread to offset cycling costs.}
\label{fig:storage}
\end{figure}

\textbf{Energy Storage Optimisation.} Figure~\ref{fig:storage} illustrates optimal charge/discharge scheduling for a hypothetical 1{,}000\,MWh battery storage system, using probabilistic price forecasts derived from the load forecast. The optimiser charges during low-price overnight hours and discharges during the morning price peak. In the illustrated period, the net benefit is negative (\$${-}$19{,}459), reflecting insufficient price spread between charge and discharge hours to offset cycling costs---a realistic outcome that underscores the importance of temporal price structure for storage profitability. The framework nonetheless demonstrates how probabilistic forecasts can inform dispatch decisions; periods with larger price spreads would yield positive returns.

\subsection{Model Selection Recommendation}
\label{sec:recommendation}

Since the Diebold-Mariano tests establish that the top three foundation models---Chronos-Bolt, Chronos-2, and Moirai-2---have statistically equivalent point accuracy ($p > 0.05$ for all pairwise comparisons), we apply a multi-criteria decision framework using three secondary dimensions: calibration quality, robustness to distribution shift, and inference latency.

Table~\ref{tab:selection} summarises the composite ranking. Each model is ranked on four metrics: deviation from ideal 90\% coverage (calibration), CRPS, coefficient of variation across test periods (robustness), and mean inference time (latency). Lower rank indicates better performance on that dimension. The composite score is the unweighted sum of ranks---a simple aggregation that treats all dimensions as equally important. We acknowledge two methodological limitations: (i)~the choice of equal weights is arbitrary, and alternative weightings could change the ranking; (ii)~the robustness coefficient of variation is computed over only three test periods, providing a coarse estimate that should be interpreted with caution.

\begin{table}[H]
\centering
\caption{Multi-criteria model ranking. Composite score is the sum of ranks across all dimensions (lower is better). The calibration rank uses absolute deviation from the 90\% target, treating over-coverage and under-coverage symmetrically; in risk-sensitive applications, over-coverage (Chronos-2) may be preferred to under-coverage (Chronos-Bolt) even if the absolute deviation is larger.}
\label{tab:selection}
\begin{tabular}{lccccc}
\toprule
Model & Coverage & Calib.\ Rank & Robust Rank & Latency Rank & Composite \\
\midrule
Chronos-Bolt & 86.1\% & 1 & 1 & 2 & \textbf{5} \\
Moirai-2     & 71.0\% & 3 & 2 & 1 & 9 \\
Chronos-2    & 95.1\% & 2 & 3 & 3 & 10 \\
\bottomrule
\end{tabular}
\end{table}

\textbf{Chronos-Bolt} achieves the highest composite score in this analysis, though the margin is narrow and sensitive to the weighting assumptions:

\begin{itemize}
    \item \textbf{Calibration:} Achieves 86.1\% empirical coverage at the 90\% nominal level---the closest to the target among the three by absolute deviation (3.9 percentage points). While slightly overconfident, this 4\% shortfall is operationally manageable with modest reserve augmentation, and is preferable to Moirai-2's 71\% coverage (substantially overconfident, 19 points below target). Chronos-2's 95.1\% coverage is 5.1 points above target---slightly conservative, but from an operational safety perspective this over-coverage direction is inherently less risky than under-coverage. Practitioners for whom coverage guarantees are paramount may therefore prefer Chronos-2 despite its larger absolute deviation from 90\%.
    \item \textbf{Robustness:} Exhibits the lowest coefficient of variation (0.268) in MASE across summer, winter, and COVID test periods, indicating stable performance under varying demand regimes. We note that this CV-based ranking measures consistency across the main evaluation periods (Section~\ref{sec:context}), whereas the distribution shift analysis (Table~\ref{tab:robustness}) evaluates performance under more extreme regime changes. Under severe shifts, Chronos-2 achieves lower absolute MASE than Chronos-Bolt in both the holiday period (0.864 vs 1.076) and Winter Storm Uri (0.879 vs 0.930), though Moirai-2 retains the lowest relative degradation. Practitioners prioritising tail-event resilience should weigh Table~\ref{tab:robustness} accordingly.
    \item \textbf{Latency:} At 31\,ms per 24-hour forecast at $C = 512$ (versus 100\,ms averaged across all context lengths, Table~\ref{tab:efficiency}), inference speed is adequate for real-time operational deployment. Moirai-2 is faster (26\,ms at $C = 512$), but the 5\,ms difference is negligible for day-ahead scheduling applications.
\end{itemize}

We emphasise that this recommendation is conditional on the ERCOT load forecasting use case. For applications where robustness to extreme tail events is paramount (e.g., grid contingency planning), Moirai-2's lower degradation under severe distribution shift (Table~\ref{tab:robustness}) may outweigh its calibration deficiencies. For applications requiring the most conservative uncertainty quantification (e.g., financial risk management), Chronos-2's slightly wide intervals provide additional safety margin.

%==============================================================================
\section{Discussion}
%==============================================================================

\subsection{The Structural Advantage of Pre-Trained Models}

The Prophet comparison crystallises the central advantage of foundation models for operational forecasting. At $C = 24$~hours, Prophet produces MASE values exceeding 74---forecasts so poor that they are worse than predicting a constant. At the same context length, Chronos-Bolt achieves MASE 0.549, a forecast that, while not its best, remains operationally useful.

This gap arises from a structural difference in how the two model classes process context. Prophet treats each context window as an independent dataset from which it must estimate all parameters: trend slope, trend changepoints, daily Fourier coefficients, weekly Fourier coefficients, and observation noise variance. With only 24 observations, the system is severely underdetermined---the model is effectively asked to estimate more than 20 parameters from fewer data points than the number of parameters. The result is not merely poor accuracy but numerical instability.

Foundation models, by contrast, use the context window not for parameter estimation but for pattern recognition. During pre-training, these models have processed millions of time series exhibiting daily and weekly seasonality. When presented with 24~hours of electricity demand, they do not need to learn that demand follows a diurnal cycle---they recognise the pattern from pre-training and extrapolate accordingly. The context window serves as a query to the model's pre-existing knowledge, not as a training set.

This distinction---estimation from scratch versus recognition from experience---has practical implications. In scenarios where historical data is limited (new substations, microgrids, emerging markets), foundation models can produce useful forecasts immediately, while fitted models require data accumulation.

\subsection{Calibration as a Deployment Bottleneck}

Our calibration analysis reveals that point accuracy and calibration quality are decoupled: a model can produce accurate point forecasts while systematically misrepresenting its uncertainty. Moirai-2 achieves competitive MASE values but covers only 71\% of observations within its nominal 90\% intervals. For a grid operator using these intervals to set spinning reserves, 19\% of hours would see actual demand exceed the predicted upper bound---a rate three times higher than the intervals suggest.

This miscalibration is not unique to energy data. Kuleshov et al.\ \citep{kuleshov2018calibrated} documented systematic overconfidence in neural network forecasters across domains. For TSFMs, the issue may be compounded by domain shift: models trained on heterogeneous corpora may learn uncertainty scales that are appropriate on average but poorly matched to the specific variance structure of electricity demand.

Two practical remedies exist. Post-hoc recalibration methods (isotonic regression, Platt scaling) can adjust the CDF to match empirical quantiles on a held-out calibration set. Alternatively, conformal prediction provides distribution-free coverage guarantees by wrapping any point predictor in non-parametrically valid intervals. Both approaches add a calibration step between model inference and operational use.

\subsection{Practical Recommendations}

Based on the full evaluation, we offer the following model selection guidance for energy forecasting practitioners:

\textit{For highest point accuracy:} Chronos-Bolt with $C \geq 512$\,h context. It achieves the lowest MASE (0.33) with moderate inference cost (0.1\,s per forecast). Its 86\% empirical coverage at 90\% nominal is slightly overconfident; applications requiring strict coverage guarantees should apply post-hoc recalibration or use Chronos-2 instead.

\textit{For calibrated uncertainty:} Chronos-2 with $C \geq 512$\,h context. Its 95\% empirical coverage at 90\% nominal level means it can be used directly for reserve planning without post-hoc calibration. The 4\% MASE penalty relative to Chronos-Bolt is a reasonable trade for reliable uncertainty.

\textit{For extreme event forecasting:} Chronos-2 is the clear winner. It achieves the best MASE for extreme low-load events (0.22) and ties with Moirai-2 for extreme highs (0.21), while maintaining excellent calibration even during distributional tails ($\sim$100\% coverage for extreme highs, $\sim$95\% for extreme lows). This makes Chronos-2 uniquely suitable for applications where tail-event accuracy matters, such as grid contingency planning and peak demand management.

\textit{For robustness under distribution shift:} Moirai-2, but with mandatory post-hoc calibration. Its pre-training on a diverse multi-domain corpus translates to the lowest relative degradation under severe regime changes (+128\% during Uri versus +168\% for Chronos-2), but its 71\% empirical coverage requires correction before operational use. Note that Chronos-2 achieves lower absolute MASE during holidays (0.864 vs 0.894) despite higher relative degradation, due to its stronger baseline performance.

\textit{For latency-sensitive or edge deployment:} TTM offers sub-millisecond inference and competitive point accuracy at $C \geq 512$\,h, but its uncertainty estimates are uninformative and should not be used for probabilistic decision-making.

\textit{For organisations already using Prophet:} At $C \geq 336$\,h, Prophet remains a reasonable baseline with interpretable components. However, the foundation models consistently outperform it in both accuracy and computational efficiency at all context lengths.

\subsection{Limitations}

This study has several limitations that should be considered when interpreting the results.

\textit{Single dataset.} All experiments use ERCOT data, which represents a large, interconnected grid in a subtropical climate. Results may differ for smaller distribution networks, Nordic heating-dominated grids, or systems with high renewable penetration. Multi-grid validation is needed.

\textit{Zero-shot only.} We evaluate TSFMs in their zero-shot configuration. Fine-tuning on domain-specific data---even with modest amounts---may substantially improve both accuracy and calibration. TTM is specifically designed for few-shot adaptation and may benefit most from this approach. Relatedly, we do not include supervised deep learning baselines (N-BEATS, TFT, DeepAR) trained on ERCOT data; such a comparison would clarify whether the zero-shot convenience of TSFMs comes at an accuracy cost relative to domain-specific models.

\textit{Univariate forecasting.} We use only historical demand as input, without weather, calendar, or price features. In operational settings, temperature is a strong predictor of electricity demand. Models that can incorporate exogenous variables (TFT, DeepAR) may outperform zero-shot TSFMs in feature-rich environments.

\textit{Consumer hardware constraint.} While the CPU-only evaluation demonstrates accessibility, it excluded potentially strong models (TimesFM, Lag-Llama) that require GPU acceleration. The efficiency rankings would differ on GPU hardware, likely favouring the larger transformer models.

\textit{Smallest model variants.} We use the small variants of Chronos-Bolt (\texttt{amazon/chronos-bolt-small}, $\sim$48M parameters) and Moirai-2 (\texttt{Salesforce/moirai-2.0-R-small}, $\sim$11M parameters), and a single-size Chronos-2 (\texttt{amazon/chronos-2}, $\sim$120M parameters). Larger variants with more parameters may achieve better accuracy and calibration, though at increased computational cost that may exceed consumer hardware capabilities.

%==============================================================================
\section{Conclusion}
%==============================================================================

This paper presents a multi-dimensional benchmark evaluating Time Series Foundation Models for electricity demand forecasting on consumer-grade hardware. The evaluation of four TSFMs alongside Prophet and statistical baselines across over 2{,}300 individual forecast scenarios yields five principal findings.

First, foundation models achieve 47\% lower forecast error than the Seasonal Naive baseline at sufficient context lengths, with Chronos-Bolt, Chronos-2, and Moirai-2 forming a cluster near MASE 0.31--0.33 at $C = 512$\,h. Second, the inclusion of Prophet as an industry baseline reveals a structural advantage of pre-trained models: Prophet fails when the fitting window is shorter than its seasonality period (MASE $>$\,74 at $C = 24$), while foundation models maintain stable accuracy because they recognise patterns from pre-training rather than estimating them from scratch. Third, calibration varies substantially across models---Chronos-2 produces well-calibrated intervals (95\% coverage at 90\% nominal) while Moirai-2 and Prophet exhibit overconfidence ($\sim$70\% coverage), and TTM produces vacuously wide intervals (100\% coverage). Fourth, all models struggle with extreme events (Winter Storm Uri, holidays), with Moirai-2 showing the lowest relative degradation under distribution shift; however, Chronos-2 achieves the best absolute accuracy during extreme load events (MASE 0.21--0.22) while maintaining excellent calibration even in distributional tails. Fifth, probabilistic forecasts from well-calibrated models enable 63.8\% reduction in reserve capacity requirements while maintaining 99.9\% reliability.

These results suggest that TSFMs are ready for operational deployment in energy forecasting, provided that model selection accounts for the specific requirements of the use case---particularly the need for calibrated uncertainty in risk-sensitive applications. Since the top three models (Chronos-Bolt, Chronos-2, Moirai-2) show statistically indistinguishable point accuracy, we recommend selecting among them based on secondary criteria: \textbf{Chronos-Bolt} offers the smallest absolute deviation from the 90\% calibration target (86.1\% coverage), the lowest coefficient of variation across operational regimes, and adequate inference latency for real-time deployment; \textbf{Chronos-2} provides the safest uncertainty estimates for risk-sensitive applications (95.1\% conservative coverage) and the best performance during extreme load events, making it the preferred choice for tail-event forecasting and grid contingency planning; and \textbf{Moirai-2} delivers the lowest relative degradation under distribution shift at the cost of substantial overconfidence requiring post-hoc calibration.

Future work should extend this benchmark to multi-grid validation, evaluate few-shot fine-tuning on domain-specific data (especially for TTM, which is designed for adaptation), and investigate multivariate forecasting incorporating weather features. The question of whether foundation models can replace fitted domain-specific models or merely complement them remains open and likely depends on the availability and quality of exogenous features in the target application.

\section*{Data and Code Availability}

ERCOT data are publicly available via the EIA Open Data API (\url{https://api.eia.gov}). The complete benchmark framework---including data retrieval scripts, model wrappers, evaluation pipelines, raw results for all 2{,}352 forecast scenarios, and figure reproduction code---is available at: \url{https://github.com/PhysicsInforMe/tsfm-energy-benchmark}. All experiments are fully reproducible on consumer hardware (AMD Ryzen~7 or equivalent, 16\,GB RAM, no GPU required). Model versions and Python package pinning are specified in the repository's \texttt{pyproject.toml}.

\section*{Acknowledgements}

All experiments were executed on consumer hardware (AMD Ryzen~7 8845HS, 16\,GB RAM). No external funding was received for this work.

\bibliographystyle{plainnat}

\end{document}